\pdfoutput=1

\documentclass[11pt]{article}

\usepackage[final]{acl}

\usepackage{times}
\usepackage{latexsym}

\usepackage[T1]{fontenc}
\usepackage[utf8]{inputenc}

\usepackage{microtype}
\usepackage{inconsolata}

\usepackage{amsmath,amssymb,amsthm,graphicx,subfigure,hyperref,algorithm,algpseudocode}
\usepackage{bbm,mathtools,caption,multirow}
\usepackage[noadjust]{cite}
\usepackage{parskip}
\usepackage{fancyvrb}

\newcommand{\bfs}{\textbf{BFS}}

\title{BFS-Prover: Scalable Best-First Tree Search for LLM-based Automatic Theorem Proving}

\author{
    \textbf{Ran Xin}\textsuperscript{1,\thanks{Key contributors.}}\quad
    \textbf{Chenguang Xi}\textsuperscript{1,*}\quad 
    \textbf{Jie Yang}\textsuperscript{2,*}\quad
    \textbf{Feng Chen} \textsuperscript{3,*,\thanks{Work done during an internship at ByteDance Seed.}} \\[0.5em]
    \textbf{Hang Wu}\textsuperscript{2}\quad 
    \textbf{Xia Xiao}\textsuperscript{1}\quad
    \textbf{Yifan Sun}\textsuperscript{1}\quad
    \textbf{Shen Zheng}\textsuperscript{1}\quad
    \textbf{Kai Shen}\textsuperscript{1}
    \\[1em]
    {\textsuperscript{1}ByteDance Seed} \ \ \
    {\textsuperscript{2}ByteDance Applied Machine Learning} \ \ \
    {\textsuperscript{3}Stanford University} \\[0.5em]
    \textbf{Correspondence:} \{ran.xin, shen.kai\}@bytedance.com
}

\date{}
\begin{document}
\maketitle

\begin{abstract}
Recent advancements in large language models (LLMs) have spurred growing interest in automatic theorem proving using Lean4, where effective tree search methods are crucial for navigating the underlying large proof search spaces. While the existing approaches primarily rely on value functions and/or Monte Carlo Tree Search (MCTS), the potential of simpler methods like Best-First Tree Search (BFS) remains underexplored. In this paper, we investigate whether BFS can achieve competitive performance in large-scale theorem proving tasks. We present \texttt{BFS-Prover}, a scalable expert iteration framework, featuring three key innovations. First, we implement strategic data filtering at each expert iteration round, excluding problems solvable via beam search node expansion to focus on harder cases. Second, we improve the sample efficiency of BFS through Direct Preference Optimization (DPO) applied to state-tactic pairs automatically annotated with compiler error feedback, refining the LLM's policy to prioritize productive expansions. Third, we employ length normalization in BFS to encourage exploration of deeper proof paths. \texttt{BFS-Prover} achieves a state-of-the-art score of $72.95\%$ on the MiniF2F test set and therefore challenges the perceived necessity of complex tree search methods, demonstrating that BFS can achieve competitive performance when properly scaled. To facilitate further research and development in this area, we have open-sourced our model at \url{https://huggingface.co/ByteDance-Seed/BFS-Prover-V1-7B}.
\end{abstract}

\section{Introduction}
\label{sec:intro}
Automatic theorem proving (ATP) in formal languages has recently become a critical benchmark for evaluating the reasoning capabilities of large language models (LLMs). By encoding mathematical problems into formal systems like Lean4, ATP enables the generation of machine-verified proofs for complex mathematical propositions, ensuring logical correctness \citep{moura2021lean,polu2020generative}. Despite the remarkable success of LLMs in natural language mathematics and reasoning tasks \citep{lewkowycz2022solving, openai2023gpt4}, theorem proving in formal languages presents unique challenges~\citep{ai4formal,autoformalization,bc-prover,lego-prover, theorem_llama, lean_star, deepseek_prover_v1}. Unlike informal reasoning, formal systems require strict adherence to syntax and semantics, as well as the ability to generate valid steps within a highly constrained formal framework. Additionally, the action (tactic) space in ATP is vast, as each proof state can lead to numerous potential tactics, making the search process for valid proofs computationally intensive \citep{polu2022formal}.

Tree search algorithms are fundamental to ATP, allowing policy models to navigate large and complex proof space efficiently \citep{polu2020generative}. Among these methods, Monte Carlo Tree Search (MCTS)~\citep{mcts} has gained popularity due to its ability to balance exploration and exploitation using value functions (critic models) or intrinsic rewards \citep{browne2012survey, silver2016alphago}. MCTS has demonstrated remarkable success in AlphaZero-style frameworks for games like chess and Go \citep{silver2017mastering}, where the underlying state spaces are vast but terminal states are well-defined. However, applying MCTS and/or value functions to ATP comes with distinct complications. While games have clear winning and losing conditions, proof search lacks such definitive end states: a proof attempt can theoretically continue indefinitely until a proof or counterexample is found, making it challenging to assess intermediate progress~\citep{han2021proofartifact, lample2022hypertree}. Additionally, ATP involves a much larger and more dynamic branching factor, sparse and delayed feedback, and an open-ended reasoning process. These differences highlight the unique demands of ATP and advocate the need for specialized adaptations of search methods to address its complexities.

Best-First Tree Search (BFS)~\citep{bfs} offers a simpler and more lightweight alternative to MCTS by prioritizing expansions based on accumulated log probabilities from the current node to the root. While its straightforward nature and computational simplicity make it attractive, BFS is often considered suboptimal for theorem proving in the existing literature~\citep{intern-prover-v2.5,hunyuanprover,deepseek-proer-v1.5} due to the following hypothesis:

\begin{itemize} 
\item Lack of efficient exploration:
BFS prioritizes high-probability paths, overlooking less likely but valid solutions. Without exploration mechanisms like upper confidence bounds or value functions, it struggles to balance exploiting promising nodes and exploring diverse paths.

\item Bias against deep reasoning paths:
BFS’s reliance on cumulative log probabilities intrinsically penalizes longer paths, as deeper expansions tend to accumulate lower scores. This bias makes BFS less effective for theorems requiring deep proofs, where intermediate states may seem unpromising but are essential for finding a solution.
\end{itemize}

\subsection{Our Contributions}
This paper challenges the prevailing perception that BFS is inherently unsuitable for large-scale ATP. We present the \texttt{BFS-Prover} system, which transforms BFS into a simple yet powerful algorithm through targeted scaling strategies. Our key contributions are as follows.

\begin{itemize}
\item \textbf{Expert iteration with self-filtering.}
We develop an expert iteration~\citep{expert-iteration} framework that strategically filters out problems solvable by beam search~\citep{beam-search} node expansions in each round. This filtering is critical as it directs training data accumulation towards harder theorems. As expert iteration progresses, the policy LLM continuously improves, learning a diverse range of tactics and deeper proofs via~BFS.

\item \textbf{Direct preference optimization (DPO) from compiler feedback.}
We use DPO~\citep{dpo} to refine the policy LLM by leveraging preference pairs naturally generated during tree search. From a given proof state, each preference pair consists of a positive tactic, which lies on the proof path, and a negative tactic, which leads to a Lean compiler error. DPO sharpens policy's distribution, enabling it to avoid unproductive tactics and thereby improving the sample efficiency of BFS.

\item \textbf{Length normalization for deeper exploration.}
We incorporate a length-normalized scoring function in BFS to mitigate its inherent bias against deeper reasoning paths. By normalizing log probabilities relative to path length, BFS can explore deeper proof paths more effectively, enabling it to solve theorems requiring long tactic chains.

\item \textbf{Empirical results on MiniF2F.}
\texttt{BFS-Prover} achieves an accumulative score of 72.95\% on the MiniF2F test set, surpassing all state-of-the-art theorem provers in the literature, including DeepSeek-Prover-V1.5~\citep{deepseek-proer-v1.5}, InternLM2.5-StepProver~\citep{intern-prover-v2.5}, and HunyuanProver~\citep{hunyuanprover}. This result underscores that \texttt{BFS-Prover} can scale to competitive levels in automated theorem proving while maintaining a lightweight design that avoids the complexities of MCTS and value functions.
\end{itemize}

\textbf{Roadmap.} The remainder of this paper is organized as follows. Section~\ref{sec:prover system} gives an overview of the \texttt{BFS-Prover} system, detailing the expert iteration framework, data filtering, DPO for policy refinement, and length normalization in BFS. Section~\ref{sec:imp} describes practical implementation and experimental results on the MiniF2F benchmark against leading prover systems. Section~\ref{sec:conclude} concludes the paper.

\section{The BFS-Prover system}\label{sec:prover system}
\begin{figure*}[htbp]
    \centering
    \includegraphics[width=0.76\textwidth]{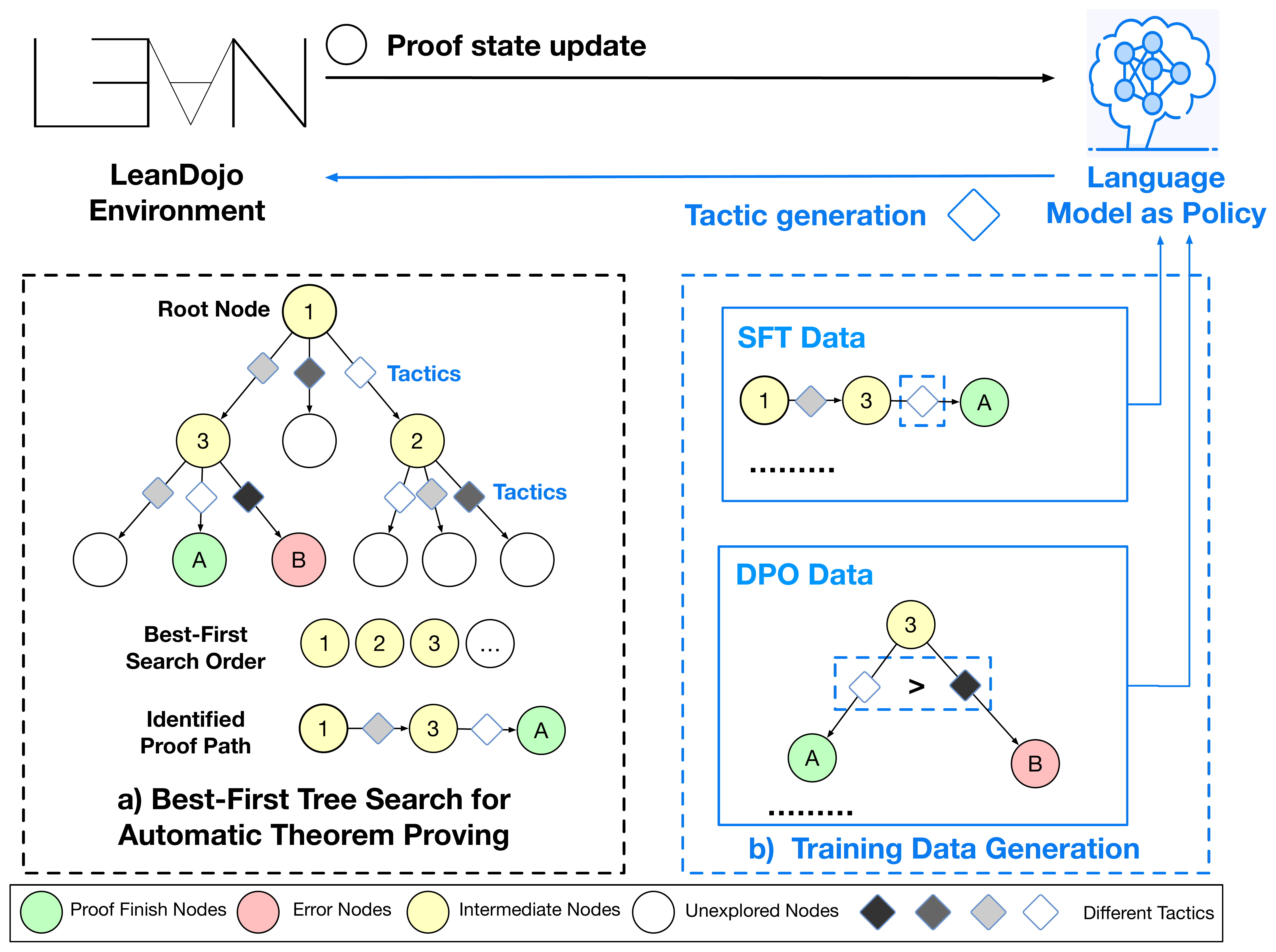}
    \caption{An overview of \texttt{BFS-Prover}. The figure illustrates two key components: a) Best-First Tree Search (BFS) mechanism for automatic theorem proving, showing how the system explores proof paths by expanding nodes based on accumulated probabilities, with different node types (proof finish, error, intermediate, and unexplored) represented via color coding. The identified proof path demonstrates a successful state-tactic sequence. b) The training data accumulation process, depicting how the system leverages both successful proofs for Supervised Fine-Tuning (SFT) and compiler error feedback for Direct Preference Optimization (DPO). The system integrates \texttt{LeanDojo} as a python interface for Lean4 and employs a language model as the policy network for tactic generation.}
    \label{fig:bootstrap}
\end{figure*}

In this section, we detail the \texttt{BFS-Prover} system design; see Fig.~\ref{fig:bootstrap} for an illustration.

\subsection{Lean4 Environment and Policy LLM}
We adopt \texttt{LeanDojo}~\citep{yang2024leandojo} as an interactive Python interface for Lean4 integration within \texttt{BFS-Prover}. It transforms Lean4 into a \texttt{gym}-like environment~\citep{gym}, facilitating interaction between the policy LLM and the formal proof assistant. Specifically, \texttt{LeanDojo} manages state transitions by executing tactics generated by the policy LLM in the Lean4 compiler. If a tactic cannot be executed, \texttt{LeanDojo} captures and returns the corresponding error messages, providing critical feedback for DPO to refine the policy LLM.

\subsection{Length-Normalized Best-First Tree Search}
\texttt{BFS-Prover} incorporates a version of Best-First Tree Search (BFS) to navigate the vast state-tactic space of proof search. The BFS engine maintains a priority queue of proof nodes (states), where the priority of each node (state) is defined
by a length-normalized scoring heuristic:
\begin{equation}\label{eq:len_norm}
\text{score}(s_L) = \frac{\sum_{t=0}^{L-1} \log p(a_t | s_t)}{L^\alpha},    
\end{equation}
where we have the following:
\begin{itemize}
    \item \(s_t\) represents the proof state at step \(t\),
    \item \(a_t\) is the tactic applied at step \(t\),
    \item $\{(s_t, a_t): t = 0, 1, ..., L-1\}$ denotes the proof path formed by applied tactics and state transitions,
    \item \(p(a_t | s_t)\) is the model’s predicted probability of applying tactic \(a_t\) at state \(s_t\),
    \item \(L\) is the total length of the path from the root to the current state \(s_L\), and
    \item \(\alpha \in [0, 1]\) is a tunable length normalization parameter.
\end{itemize}

This scoring mechanism, combined with a tunable node expansion width, enables BFS to dynamically allocate computational resources across the proof space, balancing the trade-off between exploration and exploitation. For example, increasing \(\alpha\) and/or reducing the expansion width drives the search system toward exploring deeper paths, encouraging the discovery of complex proofs that may require a long chain of tactics. 

At each node expansion step, the policy LLM generates, via a certain sampling mechanism, a list of tactics which correspond to edges in the proof tree. \texttt{LeanDojo} then executes these sampled tactics in Lean4 compiler and returns their outcomes. For each tactic application, three outcomes are possible: (1) if the tactic results in a valid proof state, a regular tree node is created and added to the node queue; (2) if the tactic completes the proof, a proof finish node is created and the proof is returned; (3) otherwise, a terminal error node is generated to represent an invalid path.

\subsection{Expert Iteration}\label{sec:ei}
\texttt{BFS-Prover} employs an expert iteration pipeline to iteratively enhance the policy LLM’s ability to navigate complex proof spaces. Given a corpus of unsolved formal statements in Lean4, each round of expert iteration consists of the following steps.

\begin{enumerate}

    \item \textbf{Beam Search Filtering:}  
    Formal statements that can be proved through BFS with beam search~node expansions are identified. These statements are then removed from the corpus and their corresponding proof data, although new, is deliberately not added to the cumulative training dataset. Beam search, being deterministic, reliably selects the most confident tactics produced by the current policy LLM. Therefore, proofs solvable under this approach are considered relatively simple, as they align closely with the strengths of the current \texttt{BFS-Prover} system. By strategically filtering out these simpler~proofs, the training data corpus is iteratively enriched with more challenging and diverse examples. This iterative refinement ensures that the policy LLM is progressively exposed to increasingly complex reasoning patterns, thereby enhancing its capability to address harder theorems in subsequent iterations.
    
    \item \textbf{Data Collection:} We then perform BFS with temperature-based sampling expansions to search for proofs of the remaining unproved formal statements in the corpus. Upon completion, the system collects all valid (proof state, tactic) pairs encountered along successful proof paths, which are subsequently added to the cumulative training dataset. The corresponding proved statements are then removed from the corpus.
    Additionally, invalid tactics that result in Lean compiler errors are recorded as on-policy negative examples to support DPO refinement.

    \item \textbf{Supervised Fine-Tuning (SFT):}  
    After each data collection phase, a new policy LLM is trained by performing SFT on a base model using the full accumulated training data corpus, which consists of all (proof state, tactic) pairs generated during previous expert iterations.     

    \item \textbf{Direct Preference Optimization (DPO):}  
    Instead of re-training the policy LLM using SFT, DPO can be applied as an alternative method to refine the current policy model by leveraging the on-policy Lean error data collected in step 2. Along the proof path, certain generated tactics, expanded from a proof state but not part of the verified proof path, result in Lean compiler errors. These erroneous tactics naturally serve as negative examples when compared to their corresponding valid proof tactics, forming preference tuples. Specifically, for a state $s$ on a proof path, we may form pairs $(a_w, a_l)$ where $a_w$ is the tactic on the proof path and $a_l$ is an error-causing tactic expanded from $s$ if exists; see~Fig.~\ref{fig:bootstrap} for an illustration. The DPO loss~$\mathcal{L}_{\text{DPO}}(\theta)$~\citep{dpo} is then defined as
    \[
    -\mathbb{E}_{(s,a_w,a_l)} \!\left[\log \sigma(\beta(r_\theta(s,a_w) - r_\theta(s,a_l)))\right]\!,
    \]
    where the differemce, i.e., implicit reward, $r_\theta(s,a) = \log(p_\theta(a|s)) - \log(p_{\text{ref}}(a|s))$ is the log probability ratio between the policy model $p_\theta$ and reference model $p_{\text{ref}}$, $\beta$ is a KL regularization parameter controlling the sharpness of the learned preferences, and $\sigma$ is the sigmoid function. By explicitly incorporating these negative signals that are absent in SFT, DPO sharpens the LLM's output distribution, thereby improving the sample efficiency and scalability of BFS within the expert iteration framework.

\end{enumerate}

This expert iteration pipeline enables the policy LLM to continually enhance its ability to generate productive tactics while adapting to progressively more challenging proof scenarios as the training data corpus grows.

\section{Practical Implementation and Benchmark Results}\label{sec:imp}
\label{sec:methodology}
In this section, we discuss practical implementation details of the \texttt{BFS-Prover} system and present its benchmark results on MiniF2F test. Throughout all experiments, we use Lean 4.7.0.

\subsection{Model, Data, and Training Setup}
\textbf{Base model and initial training data}. For illustration purposes, we use Qwen2.5-Math-7B~\citep{yang2024qwen2} as the base model for fine-tuning the policy LLM in \texttt{BFS-Prover}. In order to properly initialize the expert iteration process, we leverage proof data extracted by LeanDojo~\citep{yang2024leandojo} from Mathlib~\citep{moura2021lean}, which serves as the cold-start dataset. As expert iteration progresses, we further incorporate state-tactic datasets from Lean-Github~\citep{lean-github}, a compilation of Lean4 repositories from GitHub, and Lean-Workbook~\citep{lean-workbook}, which focuses on Olympia-level algebra and analysis. These datasets span a diverse range of mathematical topics and formal reasoning tasks, equipping the policy model with the foundational skills necessary for effective tactic generation and proof navigation.

\textbf{Formal statement corpus}. To construct our expert iteration data corpus, we autoformalize the NuminaMath-CoT dataset~\citep{li2024numinamath} using proprietary tools. We augment this with unproven theorems from Mathlib and formal statements from Lean-Workbook. The resulting corpus comprises approximately 900,000 formal mathematical statements without proofs, providing a comprehensive foundation for expert iteration.

\textbf{Training setup and procedures}. At each expert iteration round, we obtain a new policy LLM through either SFT or~DPO based on the procedures described in Section~\ref{sec:ei}. For SFT, we conduct training over 3 epochs using cosine learning rate decay, initializing at $2 \times 10^{-5}$ and decaying to $1 \times 10^{-6}$, with a global batch size of $128$. For DPO, we perform a single epoch of training with cosine learning rate decay, beginning at $5 \times 10^{-6}$ and decaying to $5 \times 10^{-7}$, utilizing a global batch size of $16$, while maintaining a KL regularization parameter of $\beta = 10$. All training is performed on A100 80GB GPUs. The selection between SFT and DPO at each expert iteration round depends on the volume of data generated in each iteration: SFT is implemented when substantial new data has been generated during the past iteration(s), whereas DPO is preferred for iterations with limited data generation, leveraging its sample efficiency through the incorporation of negative examples. 

\textbf{BFS configuration in expert iteration}. We set the length normalization parameter $\alpha$ to $0.0$ throughout, to minimize inductive bias in expert iteration. During the beam search filtering stage, we utilize a beam width of $32$ to identify easily solvable theorems. In the subsequent data collection phase, we employ temperature-based sampling with a temperature of $1.0$, nucleus sampling parameter $1.0$, and a sampling width of $2$, $4$, or $8$ to explore diverse proof paths. 

\subsection{Distributed Best-First Search Infrastructure}

To enable efficient large-scale parallel proof search, we implement a distributed system using \texttt{Ray} for distributed theorem proving across multiple machines, each equipped with $8$ A100 80GB GPUs and $128$ CPU cores. The target theorems are evenly divided across machines, with each machine running an independent proving pipeline. The system is composed of three main components:
\begin{itemize}
\item \textbf{GPU-Based Policy LLM Pool:} Each local machine deploys 8 instances of our 7B policy LLM, each powered by an asynchronous \texttt{vLLM} engine running on a dedicated A100 GPU. These instances form a shared pool that processes concurrent tactic generation requests.

\item \textbf{CPU-Based Prover Pool:} Each machine runs $96$ concurrent prover instances, reserving remaining CPU cores for standard system operations. Each prover instance executes independent BFS runs for its assigned theorems. In order to achieve balanced GPU utilization, provers distribute their requests across the policy LLM instances in a round-robin fashion based on their index modulo $8$. Each prover asynchronously interacts with its assigned policy LLMs and \texttt{LeanDojo} environments.

\item \textbf{Asynchronous interactions:} The overall distributed search system leverages \texttt{asyncio} to manage the high-concurrency workflow between provers and policy LLMs. Both the policy LLM and prover pools are implemented as \texttt{Ray} actors, enabling dynamic resource management through \texttt{Ray}'s runtime system. To ensure system responsiveness, we enforce timeout thresholds for both tactic execution (via \texttt{LeanDojo}) and model inference (via \texttt{vLLM}).
\end{itemize}
This distributed infrastructure design achieves near-linear scaling through efficient theorem distribution across machines while maximizing hardware utilization within each machine without cross-machine communication overhead.

\begin{figure*}[htbp]
\centering
\includegraphics[width=1.0\textwidth]{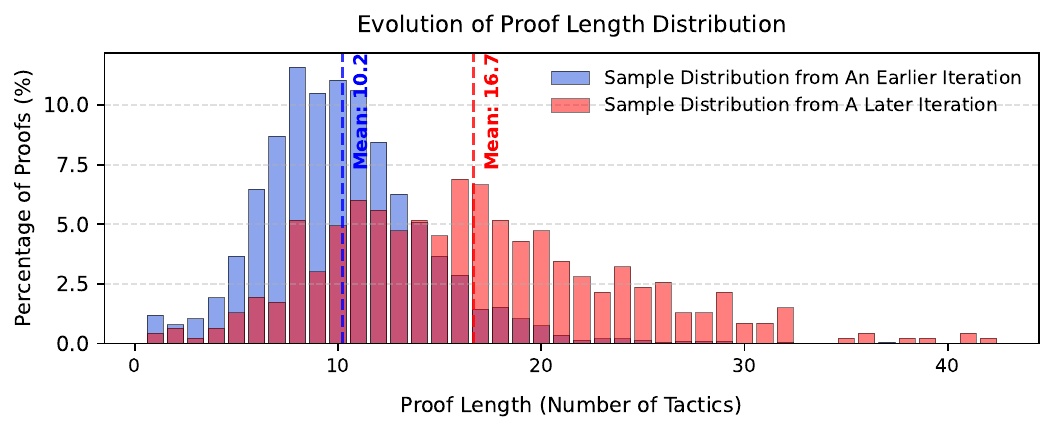}
\caption{Sample evolution of proof length distribution across expert iteration rounds. The blue distribution shows a sample distribution of proof lengths in an earlier iteration, with a mean of 10.2 tactics per proof and a concentration between 6-12 tactics. The red distribution represents a sample from a later iteration, showing a clear shift toward longer proofs with a mean of 16.7 tactics and a broader spread extending beyond 30 tactics. This shift illustrates the growing capability of \texttt{BFS-Prover} to solve increasingly complex theorems requiring deeper reasoning chains.}
\label{fig:proof_length}
\end{figure*}

\subsection{Distribution Shift in Expert Iteration}
In this subsection, we discuss and illustrate how distribution shifts emerge at both the proof level and the tactic level throughout expert iteration, shedding light on the progressively improving capabilities of \texttt{BFS-Prover} in theorem proving.

\textbf{Proof level.} A critical indicator of the effectiveness of a prover system, such as \texttt{BFS-Prover}, is its ability to discover deep proofs. We define proof length as the number of tactics found by a system to complete a proof. We observe that the distribution of proof length at each expert iteration round often exhibits a Gaussian or mixture of Gaussians, reflecting the diversity of~theorem complexities within the formal statement corpus. Intriguingly, as expert iteration progresses, the mean proof length tends to increase, indicating that BFS can find increasingly deeper and more challenging proofs as the policy LLM improves; see Fig.~\ref{fig:proof_length} for an illustration. This observation highlights the effectiveness of the expert iteration framework and the scalability of BFS, i.e., its ability to progressively tackle more complex proofs by leveraging iterative policy refinement and improved search capabilities.

\begin{figure*}[htbp]
\centering
\includegraphics[width=1.0\textwidth]{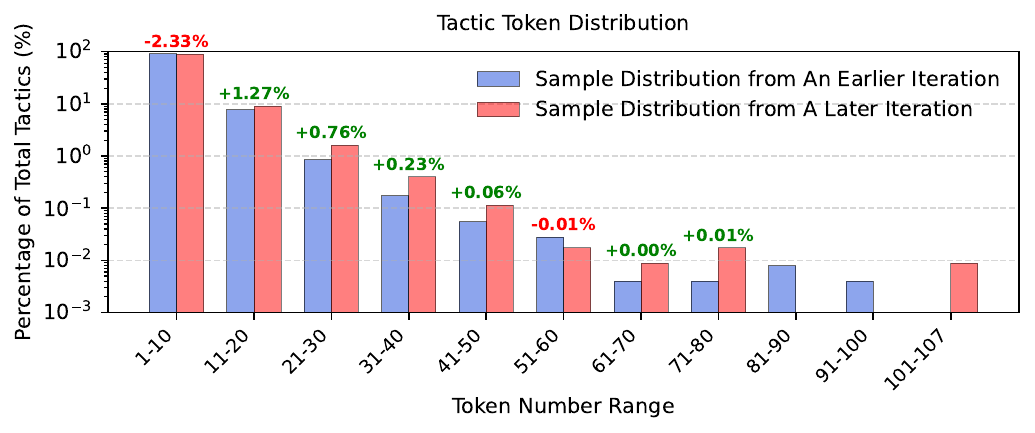}
\caption{Sample evolution of tactic token length distribution across expert iteration rounds. The plot shows the log-scaled percentage of tactics falling into different token length bins. We have two observations: (1) robustness against mode collapse during expert iteration, evidenced by the maintained diversity across token lengths, and (2) a gradual shift from an emphasis on very short tactics ($1$–$10$ tokens, decreasing by about $2.3\%$) to a higher prevalence of moderate-length tactics ($11$–$50$ tokens, with incremental increases ranging from approximately $1.27\%$ to $0.06\%$). This evolution reflects the improved ability of \texttt{BFS-Prover} to generate more complex, informative tactics while still retaining the capacity to produce concise ones when needed.}
\label{fig:token_num}
\end{figure*}

\textbf{Tactic level.} Beyond proof-level evolution, we also observe interesting distributional shifts at the tactic~level throughout expert iteration; see Fig.~\ref{fig:token_num}. Notably, the policy LLM in the \texttt{BFS-Prover} system maintains a diverse range of tactic lengths without collapsing to a narrow distribution - a common failure mode in reinforcement learning where models tend to converge to a limited set of high-reward actions~\citep{sutton2018reinforcement}. Instead, we observe a moderate but meaningful shift from very simple tactics (1-10 tokens) towards more commonly used tactical patterns (11-50 tokens). This shift suggests that through expert iteration, the policy LLM in \texttt{BFS-Prover} learns to generate more sophisticated tactics while maintaining the capability to use simpler ones when appropriate. The preservation of tactic diversity is crucial for effective theorem proving, as different proof states require different levels of tactical complexity, from simple term rewriting to complex algebraic manipulations.

\subsection{Results on MiniF2F}
In this subsection, we discuss \texttt{BFS-Prover}'s performance on the MiniF2F test benchmark~\citep{minif2f} that is a widely recognized dataset for evaluating formal mathematics systems. MiniF2F is consisted of a diverse collection of formalized competition-level mathematics problems. The policy LLM checkpoint for tactic generation used for evaluation was obtained by performing SFT on all state-tactic pairs accumulated through $10$ rounds of expert iteration pipeline of \texttt{BFS-Prover}, followed by an additional round of DPO refinement using Lean error signals as described in Section~\ref{sec:ei}. 

\subsubsection{Comparison with the State of the Art}
We now compare \texttt{BFS-Prover} developed in this work against leading theorem provers in the literature, including DeepSeek-Prover-V1.5~\citep{deepseek-proer-v1.5}, InternLM2.5-StepProver~\citep{intern-prover-v2.5}, and HunyuanProver~\citep{hunyuanprover}.

\textbf{Fixed tactic generation budget}. We first evaluate performance under a fixed tactic generation budget, defined as $K \times W \times N$, where $K$ is the total number of passes, $W$ is the expansion width, and $N$ is the maximum number of proof state expansions per pass.  This budget represents the total number of LLM inference calls required during proof search.
During BFS proof search, \texttt{BFS-Prover} uses a sampling temperature of $1.1$, an expansion width of~$2$, and a length normalization factor of $\alpha=0.5$ defined in~\eqref{eq:len_norm} to maximize the variance of explored proof paths across different search attempts. As shown in Table~\ref{tab:sota}, \texttt{BFS-Prover} achieves state-of-the-art performance under this setting without requiring either a critic model or MCTS-based search. We note that the incorporation of a critic model effectively doubles the number of inference calls since each state expansion requires both policy and value computations. 

\textbf{Accumulative results}. In order to maximize potential proof discovery, we conduct BFS searches under multiple hyperparameter configurations and aggregate the successful proofs. Specifically, we vary the BFS length normalization factor $\alpha$ defined in~\eqref{eq:len_norm} over $\{0.0, 0.5, 1.0\}$ to execute different exploration and exploitation trade-offs, while other parameter settings remain unchanged. This diverse search strategy allows \texttt{BFS-Prover} to accumulatively achieve a score of 72.95\% on the MiniF2F test, setting a new state-of-the-art result in the formal theorem proving literature. See Fig.~\ref{fig:scale} for the performance of \texttt{BFS-Prover} when the number of attempts is limited.

\begin{table*}[htbp]
   \centering
   \renewcommand{\arraystretch}{1.38}  
   \begin{tabular}{lcccc}
   \hline
   Prover System & Critic & Search & Tactic Budget & MiniF2F-test \\
   \hline
   \multicolumn{5}{l}{\textit{Whole-Proof Generation Methods}} \\
   DeepSeek-Prover-V1.5 \citep{deepseek-proer-v1.5} & No & MCTS & $32 \times 16 \times 400$ & 63.5\% \\
   \hline
   \multicolumn{5}{l}{\textit{Tactic Step Proving Methods}} \\
   InternLM2.5-StepProver \citep{intern-prover-v2.5} & Yes & BFS & $256 \times 32 \times 600$ & 65.9\% \\
   HunyuanProver~\citep{hunyuanprover} & Yes & BFS & $600 \times 8 \times 400$ & 68.4\% \\
   \textbf{BFS-Prover (this work)} & \textbf{No} & \bfs & $2048 \times 2 \times 600$ & \textbf{70.83\% $\pm$ 0.89\%} \\
   \textbf{BFS-Prover (this work)} & \textbf{No} & \bfs & accumulative & \textbf{72.95}\% \\
   \hline
   \end{tabular}
   \caption{Comparison between \texttt{BFS-Prover} and leading theorem provers in the literature on the MiniF2F test set. Each prover's performance is evaluated using its reported tactic generation budget, which represents the total number of inference calls during proof search. Note that DeepSeek-Prover-V1.5 is a whole-proof generation method and we decompose its tactic budget for ease of comparison. \texttt{BFS-Prover} achieves state-of-the-art performance without requiring a critic model (value function) and MCTS.}
   \label{tab:sota}
   \vspace{-0.2cm}
\end{table*}

\subsubsection{Scaling Law of BFS and Advantage of Negative Signals from DPO}
\begin{figure*}[htbp]
\centering
\includegraphics[width=0.6\textwidth]{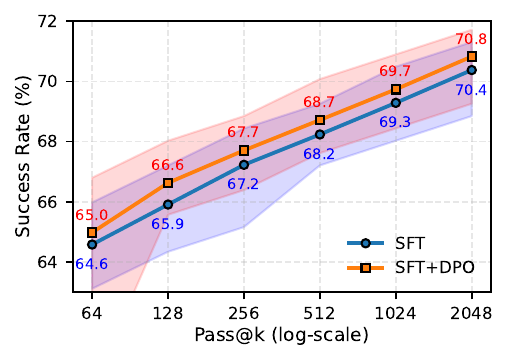}
\caption{Evolution of proof accuracies with increasing number of proof search passes on the MiniF2F test. We compare two training methods: {\color{blue}(blue)} only using SFT on all state-tactic pairs accumulated in the expert iteration pipeline of \texttt{BFS-Prover}; and {\color{red}(red)}, on top of {\color{blue}(blue)}, perform another round of DPO refinement using negative tactic examples described in Section~\ref{sec:ei}. It can be seen that DPO achieves consistent improvement in terms of both sample efficiency and overall accuracy.}
\label{fig:scale}
\end{figure*}

Finally, we investigate the search-time scaling~law of \texttt{BFS-Prover} by examining how its performance on the MiniF2F test benchmark improves with increasing proof search passes and evaluate the advantage of applying DPO to learn from negative signals in enhancing the system. We perform a total of pass@4096 and evaluate intermediate performance at pass@64, pass@128, pass@256, pass@1024, and pass@2048, where the confidence band for each intermediate pass count is computed by sampling multiple pass@64 runs. The experimental results are presented in Fig.~\ref{fig:scale}, where $x$-axis is in the log scale and the shaded region represents min-max range, i.e., the confidence band. In the following, we make several observations about key characteristics of BFS scaling in formal theorem proving. Here, SFT refers to supervised fine-tuning performed on all state-tactic pairs accumulated through the expert iteration pipeline, while SFT+DPO indicates an additional round of DPO refinement applied on top of the SFT model using on-policy negative examples from Lean4 compiler feedback as described in Section~\ref{sec:ei}. Both SFT and SFT+DPO employ the following BFS parameter configuration: a sampling temperature of $1.1$, an expansion width of~$2$, and a length normalization factor of $\alpha=0.5$.

\begin{itemize}
\item Both SFT and SFT+DPO training methods exhibit logarithmic scaling as the number of proof search passes increases. Specifically, SFT improves from 64.58\% to 70.38\% and SFT+DPO from 64.98\% to 70.83\% when scaling passes from 64 to 2048, showing consistent but diminishing returns with doubled computational budget.

\item The SFT+DPO training method consistently outperforms the SFT baseline, demonstrating the effectiveness of incorporating negative feedback from Lean4 compiler errors. This refinement enables the model to better distinguish between successful and unsuccessful proof strategies, leading to improved proof search efficiency and higher success rates.

\item Examining the min-max ranges, both methods exhibit similar variance in performance (around 3-4\% range). This indicates that while DPO improves overall success rates, it maintains comparable stability in proof search to the SFT baseline.
\end{itemize}

\section{Conclusion and Discussion}\label{sec:conclude}
This work shows that Best-First Search (BFS) can scale efficiently and achieve state-of-the-art performance in automatic theorem proving (ATP). Our results challenge the conventional belief that more complex search methods like MCTS and/or value function are necessary for large-scale formal theorem proving with LLMs. Through the development of \texttt{BFS-Prover}, we argue that a carefully designed BFS system with expert iteration which incorporates strategic data filtering, direct preference optimization, and length normalization can exceed the performance of existing approaches while maintaining computational simplicity. Our empirical results on the MiniF2F benchmark, achieving a state-of-the-art score of $72.95\%$, validate the scalability of this approach.

The success of \texttt{BFS-Prover} has several implications for the field of ATP. First, it demonstrates that algorithmic simplicity, when combined with careful scaling strategies, can outperform more complex approaches. This finding suggests that future research in ATP might benefit from focusing on refining and scaling simpler methods rather than necessarily pursuing increasingly sophisticated architectures. Second, the observed logarithmic scaling law of BFS performance with respect to computational resources suggests that while additional computation consistently yields improvements, there may be fundamental limitations to what can be achieved through increased search alone. This observation motivates future research into methods that can achieve better than logarithmic scaling.

\section*{Limitations}
While the \texttt{BFS-Prover} system demonstrates strong performance in automated theorem proving, several limitations should be acknowledged, particularly regarding model size considerations. Our current implementation relies on a relatively small 7B parameter policy model, which may limit the system's ability to learn and utilize more sophisticated mathematical reasoning patterns. While larger models (e.g., 32B or 70B parameters) could potentially capture more complex mathematical insights and generate more nuanced tactics, they would introduce significant computational challenges for both training and inference in the context of tree search. This trade-off becomes particularly evident when considering that larger models typically require more GPU memory and have longer inference latency, which could significantly reduce the number of states that can be explored within a fixed time budget. Additionally, complex mathematical proofs can generate extensive state descriptions that may exceed the practical context window of a 7B model, potentially causing the model to miss crucial information needed for generating appropriate tactics.

\bibliography{reference_xin}

\appendix

\section*{Appendix: Lean Proofs for IMO Problems Found by \texttt{BFS-Prover}}
We present the Lean 4 proofs found by \texttt{BFS-Prover} for several IMO problems in MinF2F test. These examples demonstrate that \texttt{BFS-Prover} can handle complex mathematical reasoning in formal systems, including number theory, inequalities, and geometric relationships. See Fig.~\ref{fig:imo1959p1}, \ref{fig:imo1964p2}, \ref{fig:imo1960p2}, \ref{fig:imo1962p2}, and \ref{fig:imo1983p6} for detailed proofs.

\begin{figure*}[t]
\begin{minipage}{\textwidth}
\begin{Verbatim}[frame=single,framesep=3mm,framerule=0.5mm]
theorem imo_1959_p1 (n : N)
 (h0 : 0 < n) :
 Nat.gcd (21 * n + 4) (14 * n + 3) = 1 := by {
   apply Nat.isCoprime_iff_coprime.mp
   simp [IsCoprime, h0]
   refine' ⟨-2, 3, by ring⟩
 }
\end{Verbatim}
\caption{Lean 4 proof of IMO-1959-P1 found by \texttt{BFS-Prover}.}
\label{fig:imo1959p1}
\end{minipage}
\end{figure*}

\begin{figure*}[t]
\begin{minipage}{\textwidth}
\begin{Verbatim}[frame=single,framesep=3mm,framerule=0.5mm]
theorem imo_1964_p2 (a b c : R)
 (h0 : 0 < a and 0 < b and 0 < c)
 (h1 : c < a + b)
 (h2 : b < a + c)
 (h3 : a < b + c) :
 a^2*(b + c - a) + b^2*(c + a - b) + c^2*(a + b - c) <= 3 * a * b * c := by {
   nlinarith [sq_nonneg (a - b), sq_nonneg (b - c), sq_nonneg (c - a)]
 }
\end{Verbatim}
\caption{\texttt{BFS-Prover}'s Lean 4 proof of IMO-1964-P2.}
\label{fig:imo1964p2}
\end{minipage}
\end{figure*}

\begin{figure*}[t]
\begin{minipage}{\textwidth}
\begin{Verbatim}[frame=single,framesep=3mm,framerule=0.5mm]
theorem imo_1960_p2 (x : R)
  (h0 : 0 <= 1 + 2 * x)
  (h1 : (1 - sqrt(1 + 2 * x))^2 != 0)
  (h2 : 4 * x^2 / (1 - sqrt(1 + 2 * x))^2 < 2 * x + 9) :
  -1/2 <= x and x < 45/8 := by {
  refine' ⟨_, _⟩
  all_goals field_simp at h1 h2 |-  
  all_goals rw [div_lt_iff (by positivity)] at h2
  all_goals nlinarith [
    sqrt_nonneg (1 + 2 * x), sqrt_nonneg x, 
    mul_nonneg h0 (sqrt_nonneg (1 + 2 * x)), 
    mul_nonneg h0 (sqrt_nonneg x), mul_self_sqrt h0]
  }
\end{Verbatim}
\caption{Lean 4 proof of IMO-1960-P2 found by \texttt{BFS-Prover}.}
\label{fig:imo1960p2}
\end{minipage}
\end{figure*}

\begin{figure*}[t]
\begin{minipage}{\textwidth}
\begin{Verbatim}[frame=single,framesep=3mm,framerule=0.5mm]
theorem imo_1962_p2 (x : R)
  (h0 : 0 <= 3 - x)
  (h1 : 0 <= x + 1)
  (h2 : 1/2 < sqrt(3 - x) - sqrt(x + 1)) :
  -1 <= x and x < 1 - sqrt(31)/8 := by {
    constructor
    linarith
    rw [<- sub_pos]
    field_simp [Real.sqrt_lt] at h2 |-
    apply lt_of_le_of_lt
    rw [mul_comm]
    rw [sub_eq_add_neg]
    apply lt_of_le_of_lt
    rw [<- lt_sub_iff_add_lt]
    ring_nf
    rw [<- lt_sub_iff_add_lt]
    linarith [Real.sq_sqrt (by linarith : 0 <= 1 + x)]
    rw [Real.sqrt_lt (by norm_num)]
    rw [Real.sqrt_lt] <;> nlinarith
    norm_num at this
  }
\end{Verbatim}
\caption{Lean 4 proof of IMO-1962-P2 found by \texttt{BFS-Prover}.}
\label{fig:imo1962p2}
\end{minipage}
\end{figure*}

\begin{figure*}[t]
\begin{minipage}{\textwidth}
\begin{Verbatim}[frame=single,framesep=3mm,framerule=0.5mm]
theorem imo_1983_p6 (a b c : R)
  (h0 : 0 < a and 0 < b and 0 < c)
  (h1 : c < a + b)
  (h2 : b < a + c)
  (h3 : a < b + c) :
  0 <= a^2 * b * (a - b) + b^2 * c * (b - c) + c^2 * a * (c - a) := by {
    have h1' := sq_nonneg (a - b)
    have h2' := sq_nonneg (a - c)
    have h3' := sq_nonneg (b - c)
    nlinarith [
      mul_nonneg (sub_nonneg.mpr h1.le) h1',
      mul_nonneg (sub_nonneg.mpr h2.le) h2',
      mul_nonneg (sub_nonneg.mpr h3.le) h3'
    ]
  }
\end{Verbatim}
\caption{Lean 4 proof of IMO-1983-P6 found by \texttt{BFS-Prover}.}
\label{fig:imo1983p6}
\end{minipage}
\end{figure*}

\end{document}